\documentclass[letterpaper]{article} 
\usepackage{aaai24}  
\usepackage{times}  
\usepackage{helvet}  
\usepackage{courier}  
\usepackage[hyphens]{url}  
\usepackage{graphicx} 
\usepackage{natbib}  
\usepackage{caption} 
\usepackage{algorithm}
\usepackage{algorithmic}

\usepackage{subfig}
\usepackage{epsfig}
\usepackage{graphics}
\usepackage{amsmath}
\usepackage{amssymb}
\usepackage{color, colortbl}
\usepackage{amsmath,amsfonts,mathtools,amssymb}
\usepackage{algorithm,algorithmic}
\usepackage{array}
\usepackage{color,soul}
\usepackage{xcolor}
\usepackage{bm}

\usepackage{multirow}
\usepackage{booktabs}
\usepackage{mwe}
\usepackage{enumitem}
\usepackage{tabularx}
\usepackage[super]{nth}
\newcolumntype{Y}{>{\centering\arraybackslash}X}

\DeclareMathOperator{\N}{\mathcal{N}}
\DeclareMathOperator{\I}{\mathbf{I}}
\DeclareMathOperator{\x}{\mathbf{x}}
\DeclareMathOperator{\xw}{\Tilde{\mathbf{x}}}

\def\rvx{{\mathbf{x}}}
\def\gN{{\mathcal{N}}}

\newcommand{\vect}[1]{\ensuremath{\mathbf{#1}}}

\newcommand{\xt}{\vect{x}_t}

\newcommand{\ie}{{\it i.e.~}}
\newcommand{\yy}[1]{\textcolor{black}{#1}}

\usepackage{newfloat}
\usepackage{listings}
\DeclareCaptionStyle{ruled}{labelfont=normalfont,labelsep=colon,strut=off} 
\floatstyle{ruled}
\newfloat{listing}{tb}{lst}{}
\floatname{listing}{Listing}
\title{DeS3: Adaptive Attention-Driven Self and Soft Shadow Removal \\Using ViT Similarity}
\author{
Yeying Jin\textsuperscript{\rm 1}, Wei Ye\textsuperscript{\rm 2}, Wenhan Yang\textsuperscript{\rm 3}, Yuan Yuan\textsuperscript{\rm 2},
Robby T. Tan \textsuperscript{\rm 1} \\ 
}
\affiliations{
    \textsuperscript{\rm 1}National University of Singapore\\
    \textsuperscript{\rm 2}Huawei International Pte Ltd\\
    \textsuperscript{\rm 3}Peng Cheng Laboratory\\
    e0178303@u.nus.edu, yewei10@huawei.com, yangwh@pcl.ac.cn, yuanyuan10@huawei.com, robby.tan@nus.edu.sg
}

\usepackage{bibentry}

\begin{document}

\maketitle

\begin{abstract}
Removing soft and self shadows that lack clear boundaries from a single image is still challenging. Self shadows are shadows that are cast on the object itself. Most existing methods rely on binary shadow masks, without considering the ambiguous boundaries of soft and self shadows. In this paper, we present DeS3, a method that removes hard, soft and self shadows based on adaptive attention and ViT similarity. Our novel ViT similarity loss utilizes features extracted from a pre-trained Vision Transformer. This loss helps guide the reverse sampling towards recovering scene structures. Our adaptive attention is able to differentiate shadow regions from the underlying objects, as well as shadow regions from the object casting the shadow. This capability enables DeS3 to better recover the structures of objects even when they are partially occluded by shadows. Different from existing methods that rely on constraints during the training phase, we incorporate the ViT similarity during the sampling stage. Our method outperforms state-of-the-art methods on the SRD, AISTD, LRSS, USR and UIUC datasets, removing hard, soft, and self shadows robustly. Specifically, our method outperforms the SOTA method by  16\% of the RMSE of the whole image on the LRSS dataset.
\end{abstract}

\section{Introduction}
\label{sec:intr}
Shadows can be categorized into hard shadows, soft shadows, and self shadows~\cite{salvador2004cast,huang2009moving}.
When an object blocks beams of light, shadows are formed on surfaces or nearby objects, which are referred to as hard and soft shadows. Hard shadows have sharp boundaries, while soft shadows gradually transition from shadow to non-shadow regions without any distinct boundaries.
Self shadows occur when a portion of an object obstructs beams of light, resulting in shadows being cast on the object itself.
Removing all types of shadows is intractable due to ambiguities between shadow and non-shadow regions.
Soft and self shadows are particularly challenging to identify and remove compared to hard shadows. 

\begin{figure}[t!]
	\centering
	\captionsetup[subfigure]{labelformat=empty}
	{\includegraphics[width=8.2cm]{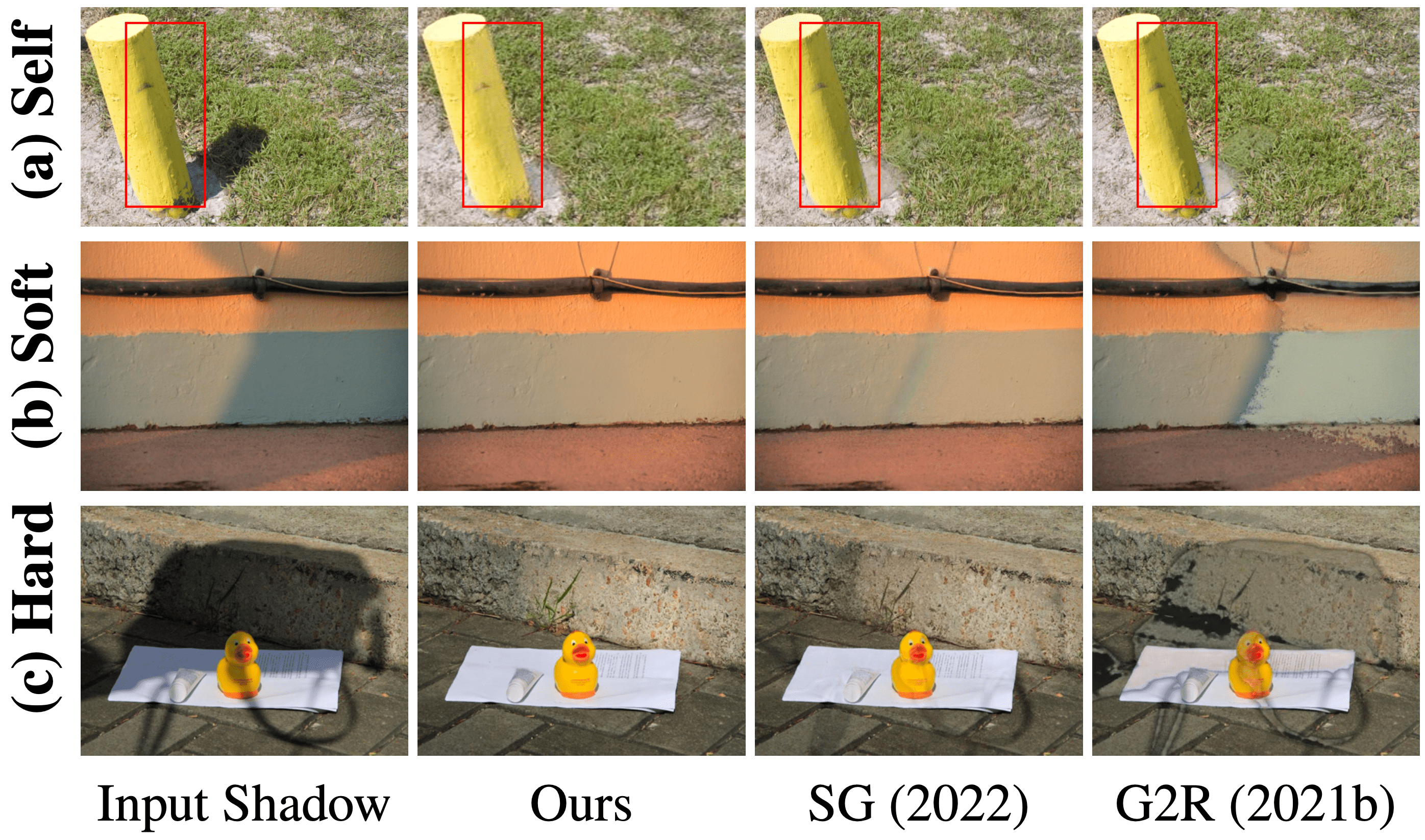}}
	\caption{The results of SOTA supervised method~\cite{wan2022sg} and weakly-supervised method~\cite{liu2021from} in removing (a) self shadow, (b) soft shadow, and (c) hard shadow. Our DeS3 can preserve meaningful objects (duck, paper, bollard, etc.) during the reverse sampling, and achieve better shadow removal results.}
	\label{fig:intro}
\end{figure}

To remove shadows, most of the methods use an off-the-shelf shadow detection method~\cite{zhu2018bidirectional} or user interaction~\cite{gryka2015learning} to obtain binary shadow masks.
However, these binary shadow masks are problematic to acquire, particularly for soft and self shadows as shown in Fig.~\ref{fig:motivation}.
Moreover, current shadow removal methods fail to handle self shadow images since obtaining the ground truth for outdoor self shadows is intractable.

The weakly-supervised methods~\cite{le2020shadow,liu2021from} require binary shadow masks to distinguish shadow and shadow-free patches.
These patches from the same images can reduce the domain gap of shadow and shadow-free domains.
However, cropping patches is time-consuming~\cite{liu2021from} and synthesizing shadows on the shadow-free regions cannot represent real-world shadow images.
Moreover, these weakly-supervised methods require shadows to be homogeneous, which is often not the case, particularly for self shadows and soft shadows.

Existing unsupervised methods~\cite{hu2019mask,liu2021shadow,jin2021dc} are GAN-based~\cite{zhu2017unpaired}, which require unpaired shadow and shadow-free images for training.
They largely rely on the statistical similarity between images from the two domains~\cite{le2020shadow} (shadow and shadow-free domains).
Unfortunately, once the two domains are statistically different, these methods produce hallucination/fake contents~\cite{zhu2018bidirectional} and suffer from unstable training.

Most SOTA methods rely on masks, but obtaining masks for self and soft shadows is intractable. 
Fixed attention in~\cite{jin2021dc} for self-shadows are unreliable.
Hence, we propose an adaptive attention to progressively focus on self-shadows.
For this reason, we employ the diffusion. Unlike other generative models, diffusion works progressively.

In this paper, we introduce {\it DeS3}\footnote{Our data and code is available at: \url{https://github.com/jinyeying/DeS3_Deshadow}.}, a diffusion-based method that removes hard, soft, and self shadows from a single image using adaptive attention and ViT similarity. 
 ``De" means  ``DeShadow",  ``S3" means three types of shadows.
Each step in the diffusion process brings about changes in our adaptive attention, as shown in Fig.~\ref{fig:adaptivea}, enabling DeS3 to effectively eliminate self and soft shadows that lack clear boundaries.
Moreover, using the ViT similarity, our DeS3 can preserve object structures even when the object is partially occluded by self and soft shadows.
To guide the reverse sampling in estimating the structural features, our basic idea is to inject the pretrained features from DINO-ViT~\cite{caron2021emerging}.
Specifically, since the keys' self-similarity~\cite{shechtman2007matching} contains object structures~\cite{tumanyan2022splicing}, they are utilized to compute the ViT similarity loss between the denoised samples and the input shadow condition.
As shown in Fig.~\ref{fig:intro}, our method can remove shadows and retain various objects (duck, paper, bollard, etc.).

In summary, we make the following contributions:
\begin{enumerate}[noitemsep,topsep=1pt]
	\item We introduce DeS3, the first shadow removal network, that performs shadow removal robustly on hard, soft and self shadows from a single image.
	\item Our DeS3 does not require shadow masks from the dataset or a shadow detector. Through the adaptive attention in the progressive diffusion process, DeS3 learns to adjust to all types of shadow regions, particularly self-shadow regions.
	\item To maintain object structure features, we integrate the ViT similarity loss into the reverse sampling. ViT's features are independent from shadows and thus can extract structures more robustly.
\end{enumerate}

Comprehensive experiments on the SRD, AISTD, LRSS, UIUC and USR datasets demonstrate that DeS3 outperforms the state-of-the-art methods, particularly on self and soft shadows.
Our DeS3 outperforms ShadowDiffusion~\cite{guo2023shadowdiffusion} by 16\% error reduction (from 3.49 to 3.01) in terms of RMSE on the overall areas of the LRSS dataset.

\section{Related Work}
\label{sec:related}
Different image priors have been explored for single image shadow removal, e.g., modeling of illumination and color~\cite{finlayson2005removal,jin2023estimating}, image regions~\cite{guo2012paired,vicente2017leave,jin2022structure}, image gradients~\cite{gryka2015learning,jin2023enhancing}.
However, traditional shadow removal methods may produce unsatisfactory results.
Recently, supervised learning-based shadow removal methods~\cite{chen2021canet,fu2021auto,liu2023decoupled,liu2023shadow} have shown promising performance.
DeshadowNet~\cite{qu2017deshadownet} removes shadows in an end-to-end manner.
DSC~\cite{hu2019direction} captures global and context information from the direction-aware spatial attention module.
SP+M-Net~\cite{le2019shadow} and SP+M+I-Net~\cite{le2021physics} remove shadow using image decomposition.
CANet~\cite{chen2021canet} is a two-stage context-aware network.
BMNet~\cite{zhu2022bijective} removes shadows using invertible neural networks.
ST-CGAN~\cite{wang2018stacked} jointly detects and removes shadows.
ARGAN~\cite{ding2019argan} uses LSTM attention to detect shadows.
DHAN~\cite{cun2020towards} employs the SMGAN to generate shadow matting.
RIS-GAN~\cite{zhang2020ris} explores the relationship of the residual images.
SG-ShadowNet~\cite{wan2022sg} treats shadow removal as intra-image style transfer.
These methods use Conditional GAN, StyleGAN~\cite{karras2019style} or Uformer~\cite{guo2023shadowformer,wang2022uformer} as their architecture.
However, supervised methods fail to handle self-shadows, as there are no ground truths (creating ones is intractable).

\begin{figure*}
	\centering
	\captionsetup[subfigure]{labelformat=empty}
	{\includegraphics[width=17.5cm]{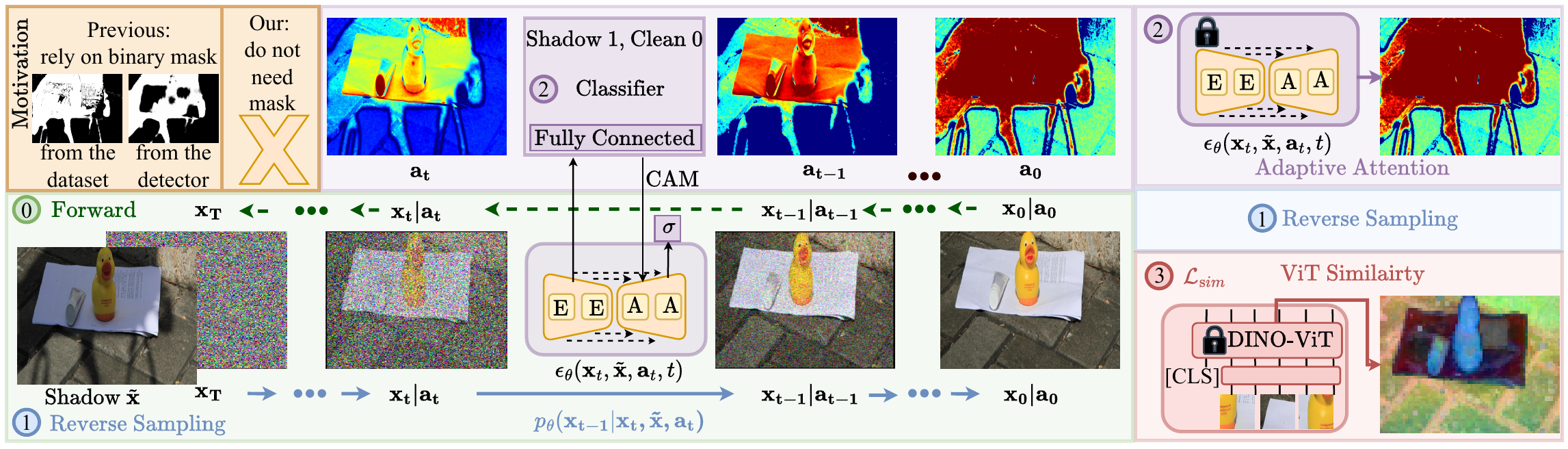}}
	\caption{The architecture and the motivation of our DeS3.
		(1) The forward diffusion is shown in green.
		The reverse sampling starts from the noise map $\x_T$ concatenated with the conditional shadow inputs $\xw$. Our DeS3 samples image $\xt$ at each time step $t$. 
		(2) We inject a classifier into the noise prediction network $\bm{\epsilon}_\theta(\x_t,\xw,\mathbf{a}_t,t)$.
		Adaptive attention $\mathbf{a}_t$ is progressively improved at each time step $t$.
		(3) To guide the reverse sampling to output the object structure features, we have the ViT similarity loss $\mathcal{L}_{\rm sim}$, extracted keys from the pre-trained DINO-ViT.}
	\label{fig:network}
\end{figure*}

The weakly-supervised shadow removal methods~\cite{le2020shadow,liu2021shadow} require shadow masks to distinguish shadow and non-shadow patches or to produce pseudo shadow pairs, making them unsuitable for handling soft or self-shadows.
Unsupervised shadow removal methods~\cite{hu2019mask,liu2021shadow,jin2021dc} need non-shadow reference images, which are hard to obtain if the input has self-shadows.
They rely on CycleGAN~\cite{zhu2017unpaired,jin2022unsupervised} and learn from unpaired data.

Denoising Diffusion Probabilistic Models (DDPM)~\cite{Ho:2020} and Denoising Diffusion Implicit Models (DDIM)~\cite{song2021ddim} have recently demonstrated promising generative ability~\cite{li2023diffusion}.
Palette~\cite{saharia2022palette}, Unit-DDPM~\cite{sasaki2021unit}, RDDM~\cite{liu2023residual} were proposed for image-to-image translation.
WeatherDiffusion~\cite{ozdenizci2023restoring} is a diffusion method presented for weather removal.
However, they are not designed to remove shadows and lack object structures~\cite{preechakul2022diffusion}.
Shadowdiffusion~\cite{guo2023shadowdiffusion} uses shadow degradation as a prior and unrolling diffusion to remove shadows. 
However, the degradation model's accuracy depends on the mask, and hence this method may not work well for soft and self-shadow removal.

\section{Proposed Method: DeS3}
\label{sec:method}

\begin{figure}[t!]
	\centering
	\captionsetup[subfigure]{labelformat=empty}
	{\includegraphics[width=0.48\textwidth]{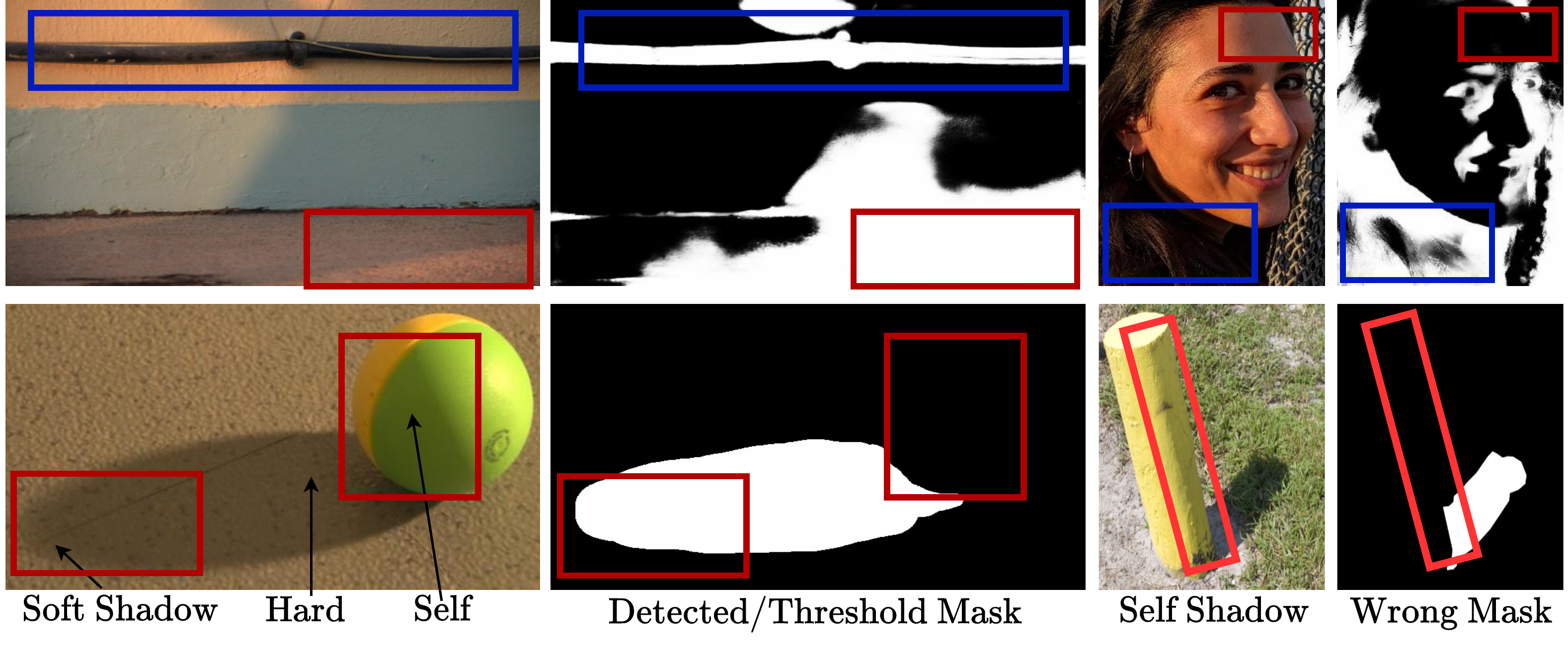}}
	\caption{Shadows can be categorized into hard, soft, and self shadows. Left: Soft and self shadows. Right: Wrong binary masks, due to the ambiguous boundaries (the red boxes for the incorrect and blue boxes for mis-detected masks).}
	\label{fig:motivation}
\end{figure}

Fig.~\ref{fig:network} shows our pipeline, comprising a forward diffusion (dashed line) and reverse sampling (solid line).
The main goal of DeS3 is to remove hard, soft and self shadows from a single image using an end-to-end network \yy(without masks).

\begin{figure}[t!]
	\centering
	\captionsetup[subfigure]{labelformat=empty}
	{\includegraphics[width=8.2cm]{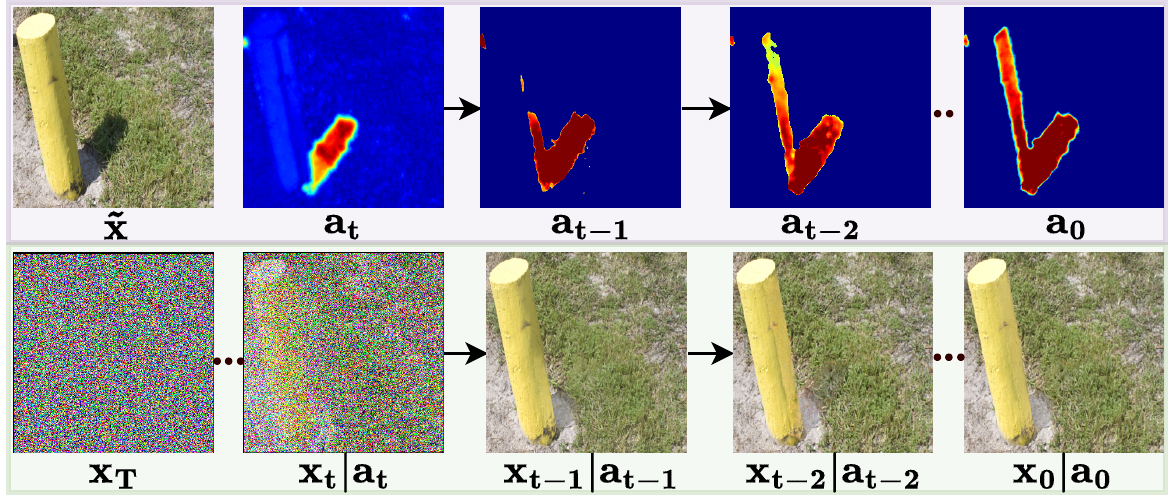}}
	\caption{Adaptive attention is refined during inference.}
	\label{fig:adaptivea}
\end{figure}

\begin{figure}[t!]
	\centering
	\captionsetup[subfigure]{labelformat=empty}
	{\includegraphics[width=8.2cm]{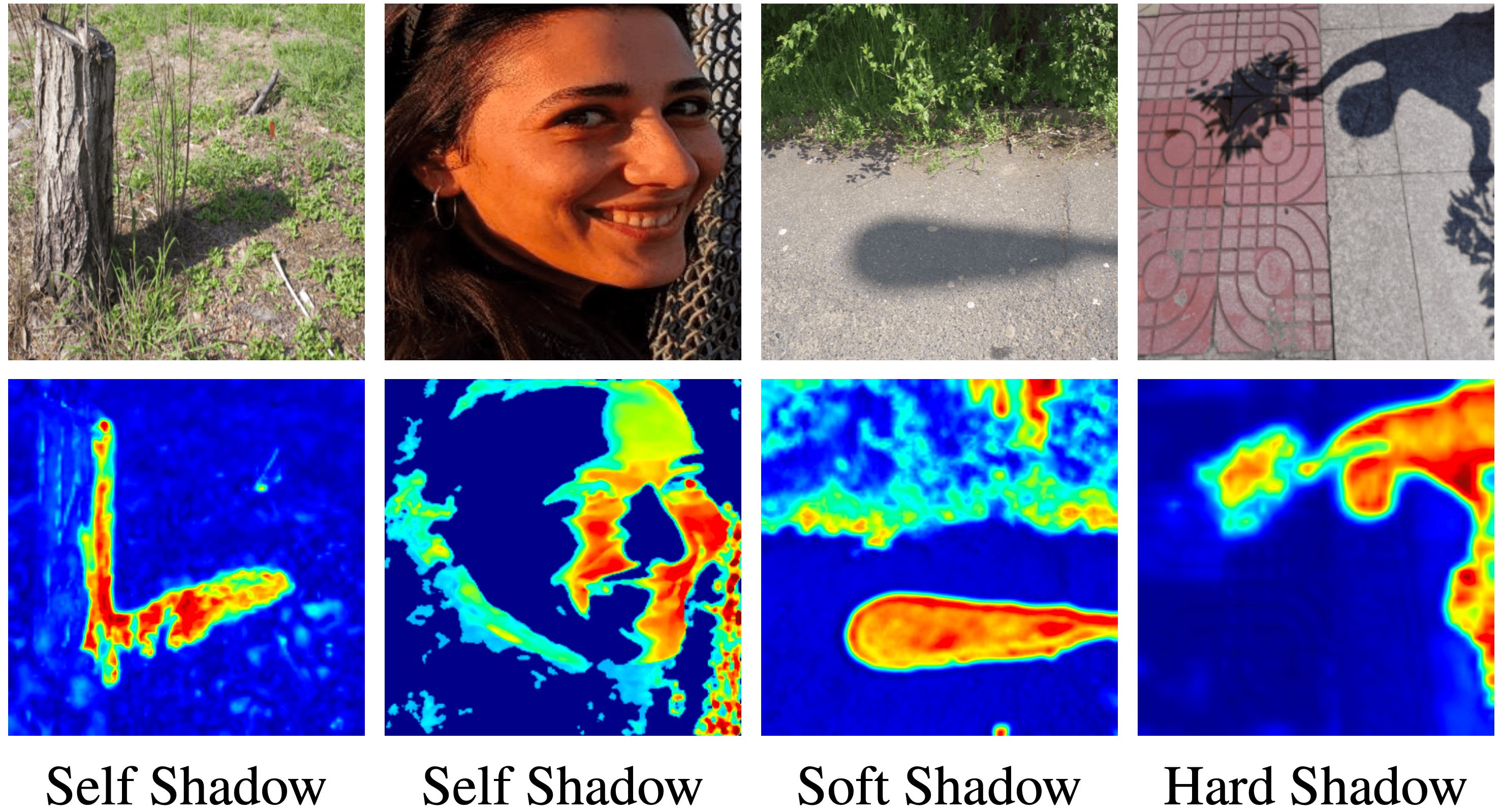}}
	\caption{The results of  adaptive attention that enables our reverse sampling to focus on hard, soft and self shadows.}
	\label{fig:attention}
\end{figure}

\begin{figure}[t!]
	\centering
	\captionsetup[subfigure]{font=small, labelformat=empty}
	\subfloat[a\label{fig:inp}]
	{\includegraphics[width = 0.162\columnwidth,height=0.8cm]{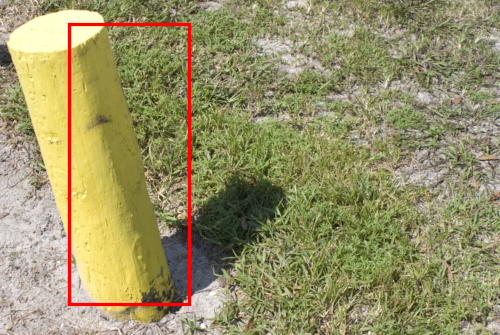}}\hfill
	\subfloat[b\label{fig:wm}]
	{\includegraphics[width = 0.162\columnwidth,height=0.8cm]{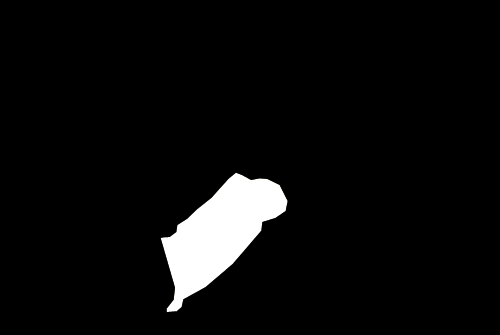}}\hfill
	\subfloat[c\label{fig:ia}]
	{\includegraphics[width = 0.162\columnwidth,height=0.8cm]{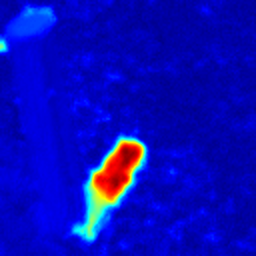}}\hfill
	\subfloat[d\label{fig:dc}]
	{\includegraphics[width = 0.162\columnwidth,height=0.8cm]{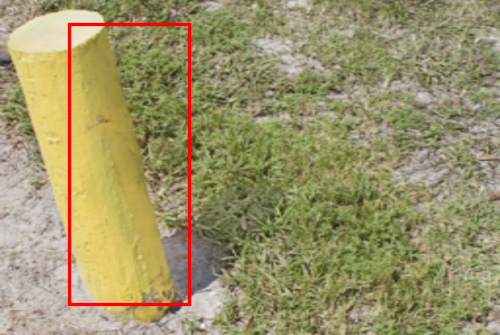}}\hfill
	\subfloat[e\label{fig:fa}]
	{\includegraphics[width = 0.162\columnwidth,height=0.8cm]{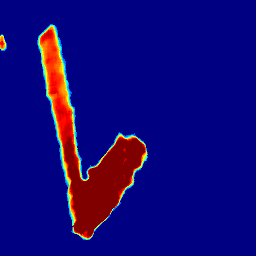}}\hfill
	\subfloat[f\label{fig:our}]
	{\includegraphics[width = 0.162\columnwidth,height=0.8cm]{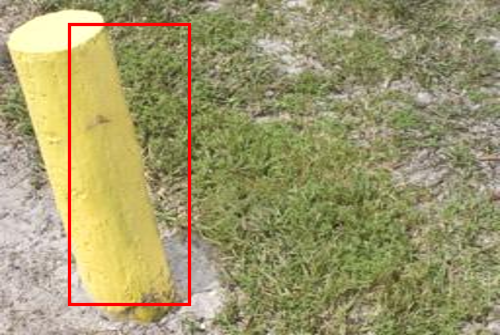}}\hfill
	\caption{d is DC~\shortcite{jin2021dc}'s result, and f is ours. We integrate adaptive attention with diffusion to tackle self-shadows.}
	\label{fig:figure_novelty}
\end{figure}

\paragraph{Conditional DDIM}
We use Denoising Diffusion Implicit Models (DDIM)~\cite{song2021ddim} as our generative model.
DDIM is trained on a clean image $\x_0 \sim q(\x_0)$ via a forward diffusion process $q(\x_t|\x_{t-1})$ that sequentially adds the Gaussian noise at every time steps $t$,
~\ie, $q(\x_{t}\vert\x_{t-1}) = \mathcal{N}(\x_{t};\sqrt{1-\beta_t}\x_{t-1},\beta_t \I)$, where $\{\beta\}_{t=0}^{T}$ is a variance schedule. The forward diffusion of length T expressed as:
\begin{align}
	\label{eq:forward}
	q(\x_{1:T}\vert\x_0) = \prod_{t=1}^T q(\x_{t}\vert\x_{t-1}).
\end{align}

Diffusion learns to reverse the process in Eq.\eqref{eq:forward} with the Gaussian transitions,~\ie, $p_{\theta}(\x_{t-1}\vert\x_t) = \mathcal{N}(\x_{t-1};\bm{\mu}_{\theta}(\x_t,t),\mathbf{\Sigma}_{\theta}(\x_t,t))$, at each time step. 

The reverse denoising process is parameterized by a trainable network (e.g., U-Net), which estimates the mean $\bm{\mu}_{\theta}(\x_t,t)$ and variance $\mathbf{\Sigma}_{\theta}(\x_t,t)$ with parameter $\theta$.
This reverse process starts from a standard normal distribution $p(\x_T)=\mathcal{N}(\x_T;\mathbf{0},\I)$ and follows: 
$\label{eq:reverse}
p_{\theta}(\x_{0:T}) = p(\x_{T}) \prod_{t=1}^T p_{\theta}(\x_{t-1}\vert\x_t).$
The noise prediction network $\bm{\epsilon}_{\theta}(\x_t,t)$ can be optimized by this objective function:
$\label{eq:objective}
L_\theta(\text{DM}) = \mathbb{E}_{\x_0,\bm{\epsilon},t}\Big{[}\|\bm{\epsilon}-\bm{\epsilon}_\theta(\x_t,t)\|^2\Big{]}.$
After the optimization, we can sample from the learned parameterized Gaussian transitions $p_{\theta}(\x_{t-1}\vert\x_t)$ by:
$\label{eq:ddim}
\rvx_{t-1} = \bm{\mu}_{\theta}(\rvx_t, t) +\mathbf{\Sigma}_{\theta}^{1/2}(\rvx_t, t)\bm{\epsilon},~~~~\bm{\epsilon} \sim \gN(\mathbf{0}, \mathbf{I}).
$
Our DeS3 use conditional DDIM to output deterministic and consistent samples.

We reverse process $p_{\theta}(\x_{0:T}|\xw)$,
without changing the diffusion process $q(\x_{1:T}|\x_0)$ in Eq.\eqref{eq:forward}.
We replace the noise prediction network $\bm{\epsilon}_{\theta}(\x_t,t)$ with
$\bm{\epsilon}_\theta(\x_t,\xw,t)$, where $\xw$ denotes the shadow image.
Here, we use the conditional DDIM~\cite{song2021ddim} and input shadow images $\xw$ and shadow-free $\x_0$ pairs $(\xw,\x_0)$ to output shadow-free image.
We concatenate $\xw$ and $\xt$ channel-wisely, and input them to the deterministic reverse process.
Therefore, we treat shadow removal as a reverse process with the loss function $L_\theta(\text{CDM})$ and parameterize a noise prediction network $\bm{\epsilon}_\theta(\x_t,\xw,t)$:
\begin{align}\label{eq:objective_c}
	L_\theta(\text{CDM})&=\mathbb{E}_{\x_0,\bm{\epsilon},t}\Big{[}\|\bm{\epsilon}-\bm{\epsilon}_\theta(\x_t,\xw,t)\|^2\Big{]},\\
	\hat\x_0(\x_{t})& = \frac{1}{\sqrt{{\bar{\alpha}_t}}}\left( 
	\xt - \sqrt{1 - {\bar{\alpha}_t}} \cdot \vect\bm{\epsilon}_\theta(\x_t,\xw,t)
	\right),
	\nonumber
\end{align}
where $\hat\x_0(\x_{t})$ refers to the estimated clean image from the sample $\x_t$. For brevity, we short $\hat\x_0(\x_{t})$ as $\x$.

The shadow removal starts from the noise map, $\x_T\sim\N(\mathbf{0},\I)$, with the condition shadow input $\xw$, and applies the diffusion towards the target non-shadow image $\x_0$.
The sampling from the reverse process $\x_{t-1}\sim p_{\theta}(\x_{t-1}\vert\x_t,\xw)$ employs:
$\x_{t-1} =  \sqrt{\bar{\alpha}_{t-1}}\left(\frac{\x_t-\sqrt{1-\bar{\alpha}_t}\cdot\bm{\epsilon}_{\theta}(\x_t,\xw,t)}{\sqrt{\bar{\alpha}_t}}\right) + \sqrt{1-\bar{\alpha}_{t-1}}\cdot\bm{\epsilon}_{\theta}(\x_t,\xw,t).
\label{eq:cddim}$

\subsection{Adaptive Classifier-driven Attention}
Our novelty lies in adaptive attention, which can highlight the regions of all three types of shadows, particularly self-shadows. 
Figs.~\ref{fig:fa} to \ref{fig:our} show our results using adaptive attention, and additional results can be found in Fig.~\ref{fig:attention}. 
Our adaptive attention is progressively refined throughout the reverse process, as illustrated in Fig.~\ref{fig:adaptivea}. 
In contrast, \cite{jin2021dc} relies solely on fixed CAMs, which tend to struggle with handling self-shadows, as demonstrated in Figs.~\ref{fig:ia} to \ref{fig:dc}.

Since a plain diffusion model is not aware of the shadow regions,
we design an adaptive attention to allow the reverse sampling to focus on the shadow regions. 
Our soft attention map differs from existing binary masks shown in Fig.~\ref{fig:motivation}, which are hard to obtain for self and soft shadows.
We fuse the attention into diffusion by injecting the classifier in the U-Net.
Specifically, by labeling the clear image $\x_0$ as 0, and the shadow condition $\xw$ as 1, and using binary classification with Class Activation Map (CAM)~\cite{Zhou_2016_CVPR}. Given an image $x \in \{\xw, \x_0\}$, $L_\text{cam}=-(\mathbb{E}_{x \sim \xw}[log(C(x))]+\mathbb{E}_{x \sim \x_0}[log(1-C(x))])$, where $C$ is the classifier. 
By leveraging the information from the classifier, the CAM attention is learned from the data and can focus on hard, soft, and self-shadow regions, as shown in Fig.~\ref{fig:attention}.

Unlike the fixed attention in~\cite{jin2021dc}, our attention is adaptive and progressively refines throughout the diffusion process. 
Besides the CAM attention, we employ a residual map refinement. 
By utilizing shadow and shadow-free pairs during training, we can employ a residual map to refine classifier-driven attention progressively.
That is to say, the noise estimation network $\bm{\epsilon}_{\theta}(\x_t,\xw,\mathbf{a}_t,t)$ has two tasks: estimating the noise and progressively refining the attention $\mathbf{a}_t$.
We add one Sigmoid Conv layer after the last layer of the noise estimation network, and obtain the residual map $\mathbf{m}_\text{res}$ by computing the difference map between shadow and shadow-free pair. This difference map is then used to guide the refinement of $\mathbf{a}_t$ via the following loss:
\begin{align}\label{eq:objective_att}
	L_\text{att}&=\mathbb{E}_{t \sim [1,T]}\Big{[}\| \mathbf{a}_t - \mathbf{m}_\text{res}  \|^2\Big{]},
\end{align}
where  $\mathbf{m}_\text{res}= \sigma(\x_0 - \xw)$, $\sigma(\cdot)$ is the Sigmoid Conv layer.

\begin{figure}[t!]
	\centering
	\captionsetup[subfigure]{labelformat=empty}
	{\includegraphics[width=8.2cm]{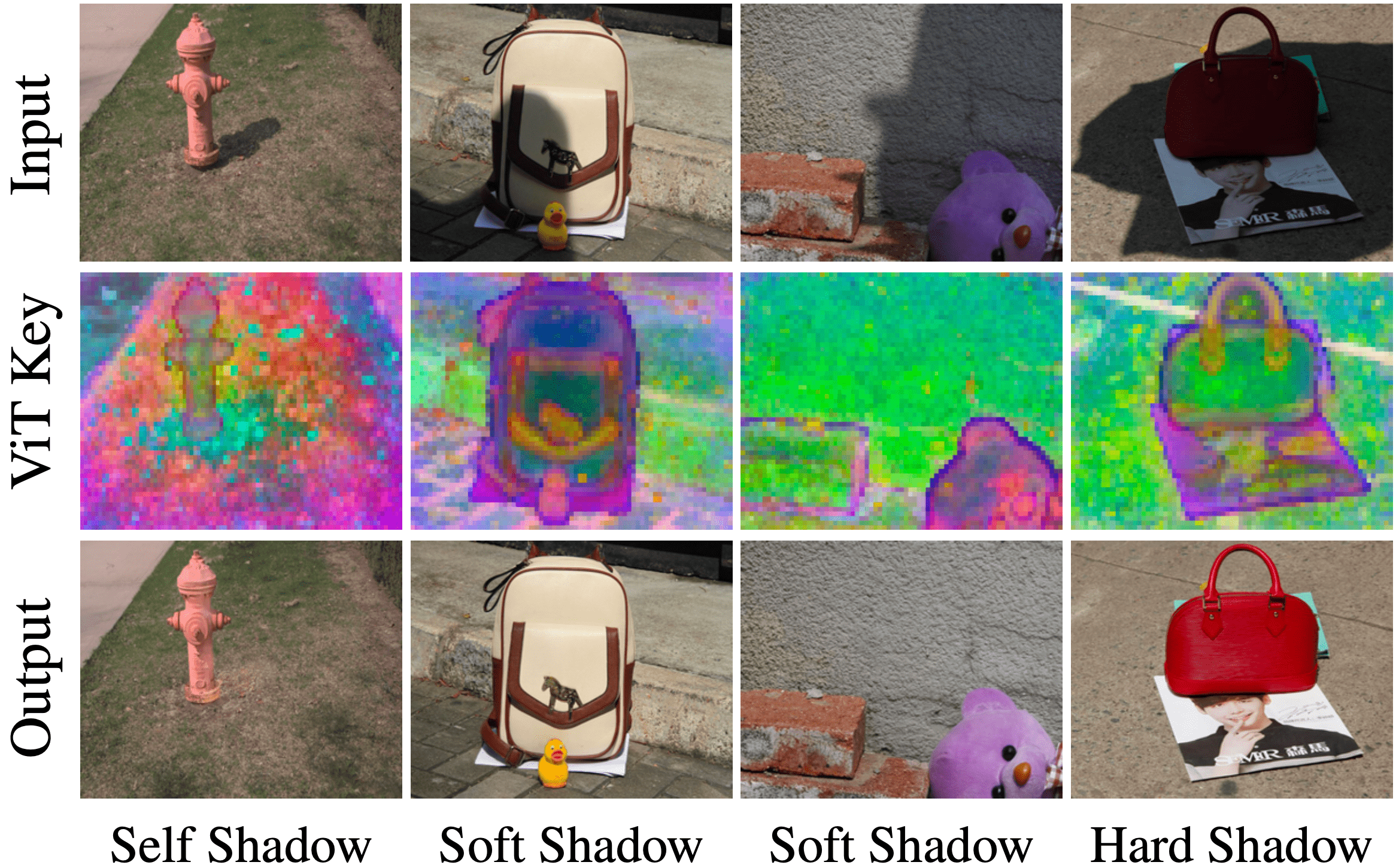}}
	\caption{We visualize the deep DINO-ViT features. These object structure ViT features help DeS3 preserve the object/scene structures (e.g., fire hydrant in self shadow example, duck, bag, bear in soft shadow examples).}
	\label{fig:vit}
\end{figure}

\begin{figure}[t]
	\centering
	\captionsetup[subfigure]{labelformat=empty}
	{\includegraphics[width=8.2cm]{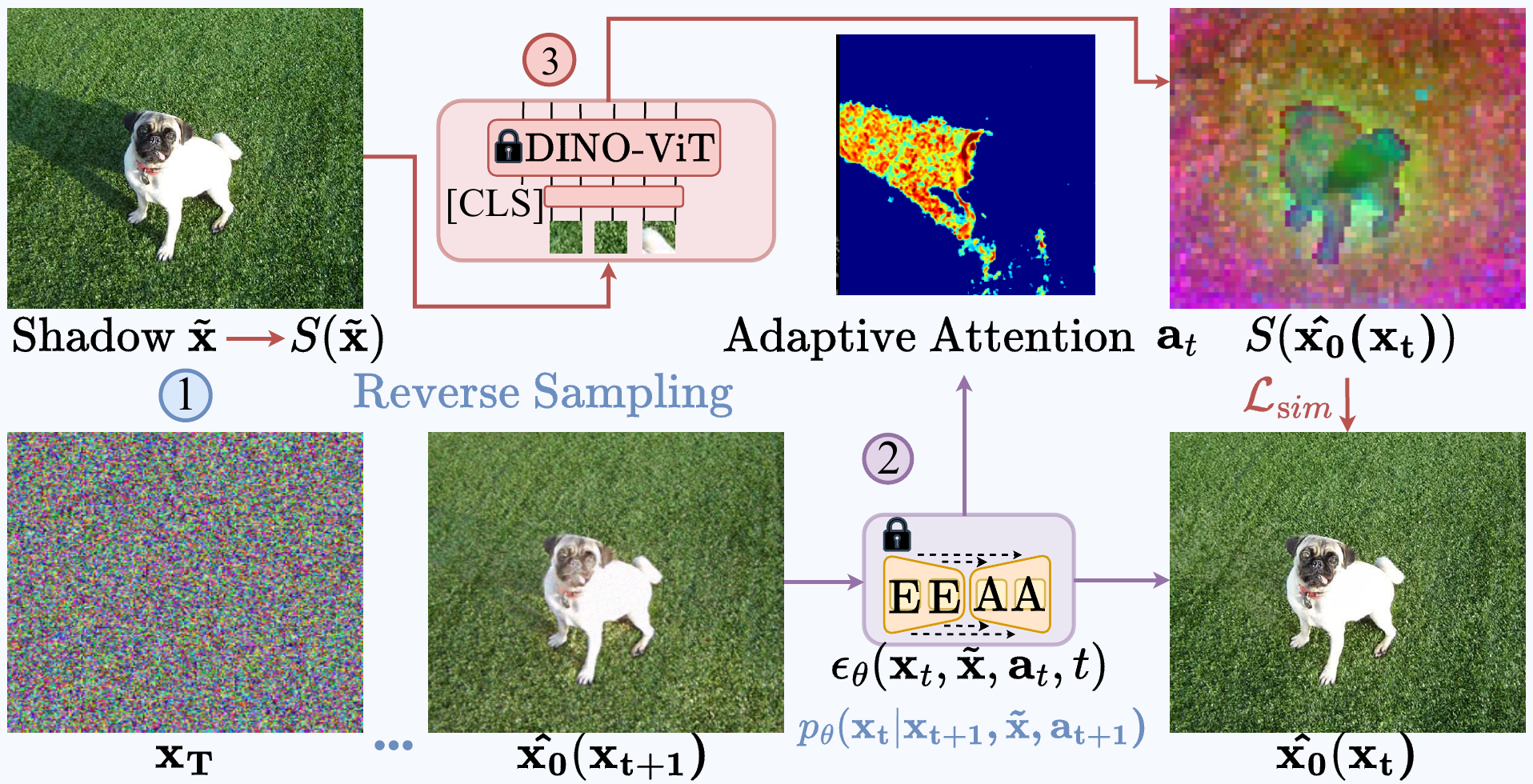}}
	\caption{In our reverse sampling process, we use the loss function to guide the output to follow the ViT similarity. \yy{When $\mathcal{L}_{\text sim}$ is high, continue the reverse sampling till 25 steps. When $\mathcal{L}_{\text sim}$ is low, the reverse sampling can be stopped early.}}
	\label{fig:chself}
\end{figure}

\begin{table*}[t!]
	\centering
	\resizebox{0.785\textwidth}{!}{
		\centering
		\begin{tabular}{l|c|ccc|ccc|ccc}
			\toprule
			\multicolumn{1}{l|}{\multirow{2}{*}{Method}} &\multicolumn{1}{c|}{\multirow{2}{*}{Train}} & \multicolumn{3}{c|}{RMSE $\downarrow$} & \multicolumn{3}{c|}{PSNR $\uparrow$} & \multicolumn{3}{c}{SSIM $\uparrow$} \\ \cline{3-11} 
			~ & & S & NS & ALL & S & NS & ALL & S & NS & ALL\\\cline{2-10} 
			\hline			
			DHAN~\shortcite{cun2020towards}    &P+M   & 8.94 & 4.80 & 5.67 & 33.67 & 34.79 & 30.51 & 0.978 & 0.979 & 0.949 \\
			Auto~\shortcite{fu2021auto} &P+M & 8.56 & 5.75 & 6.51 & 32.26 & 31.87 & 28.40 & 0.966 & 0.945 & 0.893 \\
			EM~\shortcite{zhu2022efficient} &P+M &10.00 & 6.04 &7.20  & 29.44 & 26.67 & 24.16 & 0.937 & 0.879 & 0.779 \\
			BM~\shortcite{zhu2022bijective} &P+M &6.61 & 3.61 & 4.46 &35.05 &36.02 &31.69 &0.981 & 0.982 & 0.956 \\
			SG~\shortcite{wan2022sg}  &P+M & 7.53 &3.08 &4.33 & 33.88 &36.43 & 31.38 &0.981 &0.987 &0.959\\\hline
			DC~\shortcite{jin2021dc}  &UP & 7.70 & 3.65 & 4.66 & 34.00 & 35.53 & 31.53 & 0.975 & 0.981 & 0.955 \\\hline
			SR3~\shortcite{saharia2022image} &P   & - & - & - &35.44 &34.35 & 31.29 & 0.980 & 0.970 & 0.946\\
			W.Dif.~\shortcite{ozdenizci2023restoring}&P   & - & - & - &33.38 &31.15 & 28.45 & 0.981 & 0.972 & 0.951\\\hline
			DSC~\shortcite{hu2019direction}   &P   & 8.81 & 4.41 & 5.71 & 30.65 & 31.94 & 27.76 & 0.960 & 0.965 & 0.903 \\
			Ours &P &\textbf{5.88} & \textbf{2.83} & \textbf{3.72} &\textbf{37.45} &\textbf{38.12} &\textbf{34.11} &\textbf{0.984} &\textbf{0.988} &\textbf{0.968}\\
			\bottomrule
	\end{tabular}}
	\caption{Quantitative results on the SRD dataset. S, NS and ALL represent shadow, non-shadow and entire regions. M shows ground truth shadow masks are also used in training. ``P" and ``UP" stand for ``Paired" and ``Unpaired".}
	\label{tb:srd}
\end{table*}

\begin{figure*}[t!]
	\centering
	{\includegraphics[width=\textwidth]{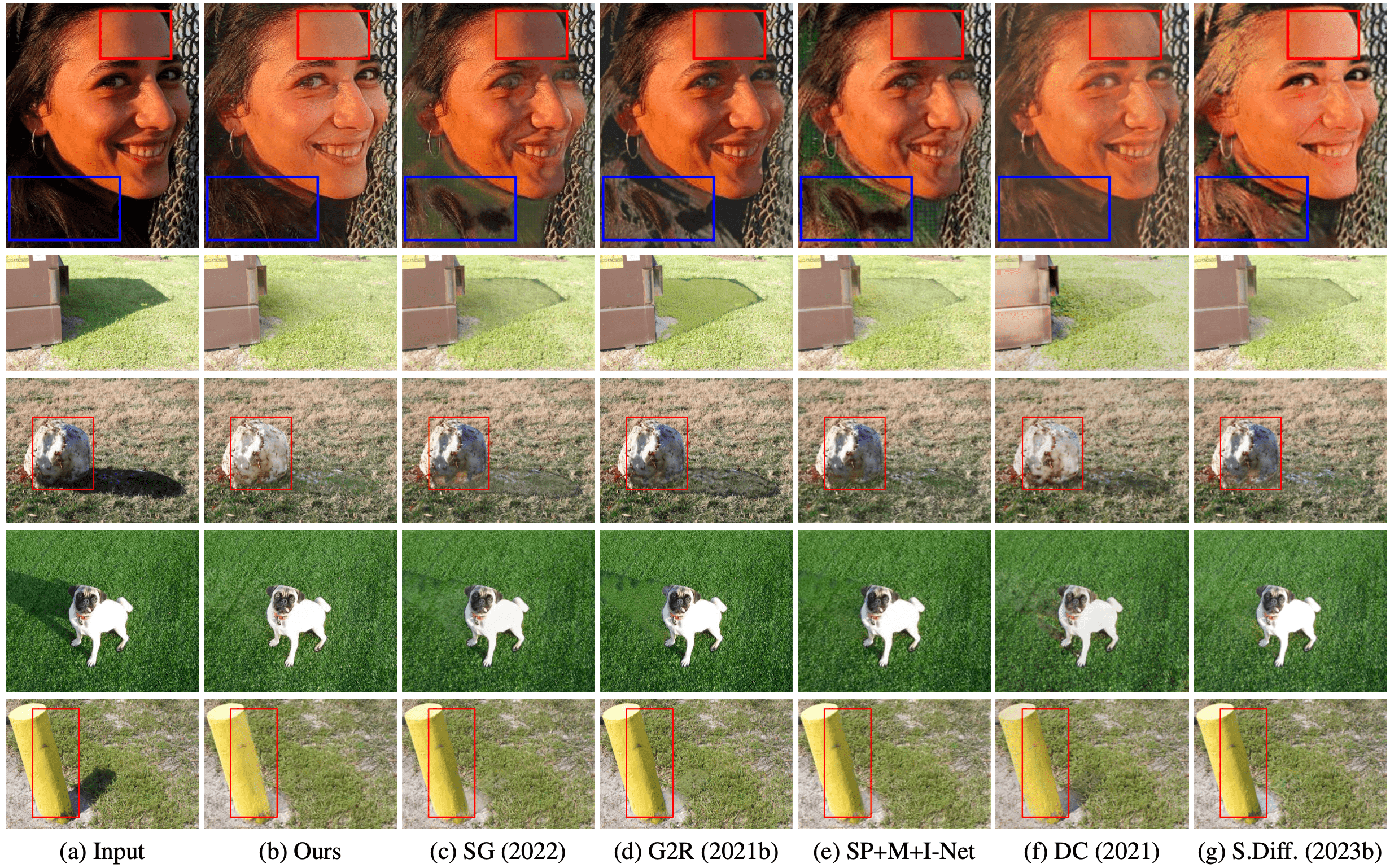}}
	\caption{Self shadow removal results on UIUC dataset. (a) Input images. (b) Our results. (c)$\sim$(g) Results of the state-of-the-art methods. Note that our method does not need additional masks in training and testing.}
	\label{fig:self}
\end{figure*}

\subsection{Object Structures in Reverse Sampling}
\begin{figure}[t!]
	\centering
	\captionsetup[subfigure]{labelformat=empty}
	{\includegraphics[width=8.2cm]{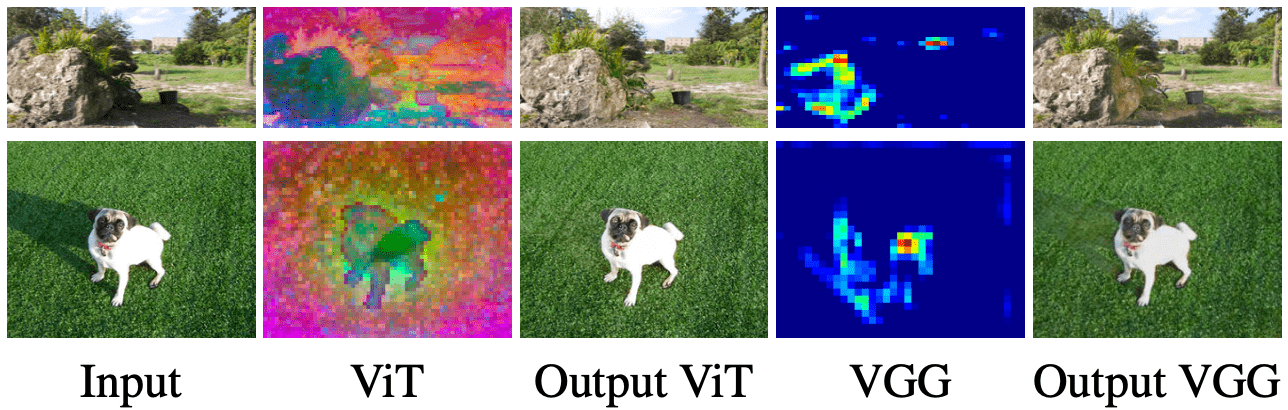}}
	\caption{Ablation study on ViT vs VGG, the output of ViT is better than VGG.}
	\label{fig:figure_feature}
\end{figure}

Unlike the VGG loss \cite{jin2021dc} used in training, we propose to use a DINO-ViT loss as a stopping criterion in the reverse sampling (inference) stage. 
DINO-ViT demonstrates greater robustness in preserving object structures across all types of shadows when compared to VGG, as shown in Fig.~\ref{fig:figure_feature}.
Moreover, in contrast to the losses  in~\cite{jin2021dc}, our DINO-ViT loss is exclusively employed during inference, serving as a stopping criterion, as depicted in Fig.~\ref{fig:chself}. If the total loss $L_t$ is smaller than that in the subsequent diffusion step, we halt the reverse process earlier.

In contrast to the existing shadow removal methods that use the supervised losses (requiring ground truth shadow-free) in training, we use the self-tuned losses in the reverse sampling (inference).
Preserving object/scene structures is critical for shadow removal~\cite{qu2017deshadownet}.
Therefore, we introduce the ViT similarity loss to guide the reverse sampling.
To preserve the object structures, our key idea is to exploit a self-similarity prior~\cite{shechtman2007matching}.
For the soft shadow example in Fig.~\ref{fig:vit}, the left and right parts of the bag contain different lighting conditions and appearances, but they do share similar structures (i.e., self-similarity).
A pre-trained DINO-ViT~\cite{caron2021emerging} can provide better object structure features, compared to the CNN backbone, transformer-based backbones shown in~\cite{karmali2022hierarchical}.
Motivated by this, we use a pre-trained DINO-ViT as our feature extractor, enabling us to capture deep features~\cite{amir2021deep}, and providing useful feature supervision to our DeS3. 

\paragraph{DINO-ViT Keys}
We extract keys (deep features) from the pre-trained DINO-ViT.
In the multi-head self-attention layer, the keys contain the object structure information~\cite{tumanyan2022splicing}.
In Fig.~\ref{fig:vit}, we show the Principal Component Analysis (PCA) visualization of the keys' self-similarity and demonstrate the top three components as RGB at the deepest layer of DINO-ViT. 
As one can observe, the feature representations capture the object structures, helping our DeS3 preserve and recover meaningful features.

We propose the ViT similarity loss between the intermediate keys of the input image and the estimated denoised output image during the reverse sampling given a noise estimation network $\bm{\epsilon}_{\theta}(\x_t,\xw,\mathbf{a}_t,t)$ and ViT model: 
\begin{align}\label{eq:loss_semantic}
	\mathcal{L}_{\text sim} (\mathbf{\xw},\mathbf{\x}) = \left \|  S^l(\mathbf{\xw})-S^l(\mathbf{\x}) \right \| _{F},
\end{align}
where $S$ is the self-similarity descriptor,
$S^{l}(x) \in \mathbb{R}^{(n)\!\times\!(n)}$, 
with $n \times n$ dimension, where $n$ is the number of patches. 
In DINO-ViT, the input image is split into $n$ non-overlapping patches.
$l=11$ is the deepest ViT layer.
$\left \| \cdot \right \| _{F}$ is the Frobenius norm.
The self-similarity descriptor is defined as:
\begin{align}\label{eq:loss_sel}
	S^l(\mathbf{\x})_{ij} = \text{cos}(k^{l}_i(\mathbf{\x}), k^{l}_j(\mathbf{\x})) = 1 - \frac{k^{l}_i(\mathbf{\x}) \cdot k^{l}_j(\mathbf{\x})} {\left \| k^{l}_i(\mathbf{\x}) \right \| \cdot \left \| k^{l}_j(\mathbf{\x}) \right \| },
	\nonumber
\end{align}
where $\text{cos}(\cdot)$ is the cosine similarity between spatial keys $k_i$ and $k_j$.

\begin{table*}
	\centering
	\renewcommand{\arraystretch}{1.2}
	\resizebox{0.70\textwidth}{!}{
		\centering
		\begin{tabular}{l|c|c|c|c|c|c|c}
			\toprule 
			Aspects &Ours &S.Diff. &SG &G2R &SP+M+I &DC &SP+M \\\hline 
			1.DeS$\uparrow$  &\textbf{8.6$\pm$1.3} &5.8$\pm$3.9&\underline{6.2$\pm$2.7} &5.6$\pm$3.8 &6.0$\pm$4.0 &5.4$\pm$1.8 &5.6$\pm$2.3 \\\hline
			2.Real$\uparrow$   &\textbf{9.0$\pm$0.9} &6.9$\pm$2.8 &6.9$\pm$2.0 &\underline{7.2$\pm$3.2} &7.0$\pm$1.5 &6.0$\pm$0.5 &6.6$\pm$1.8 \\
			\bottomrule
	\end{tabular}}
	\caption{User study on the self shadow removal of the UIUC dataset~\cite{guo2011single}, our method obtained the highest mean (the max score is 10), showing our method is effective in shadow removal (DeS) and visually realistic (Real). The best results are in bold, the second best are in underline.}
	\label{tb:tb_user}
\end{table*}

\section{Experiments}
\label{sec:experiments}
\begin{figure}[t]
	\centering
	\captionsetup[subfigure]{font=small, labelformat=empty}
	{\includegraphics[width=8.2cm]{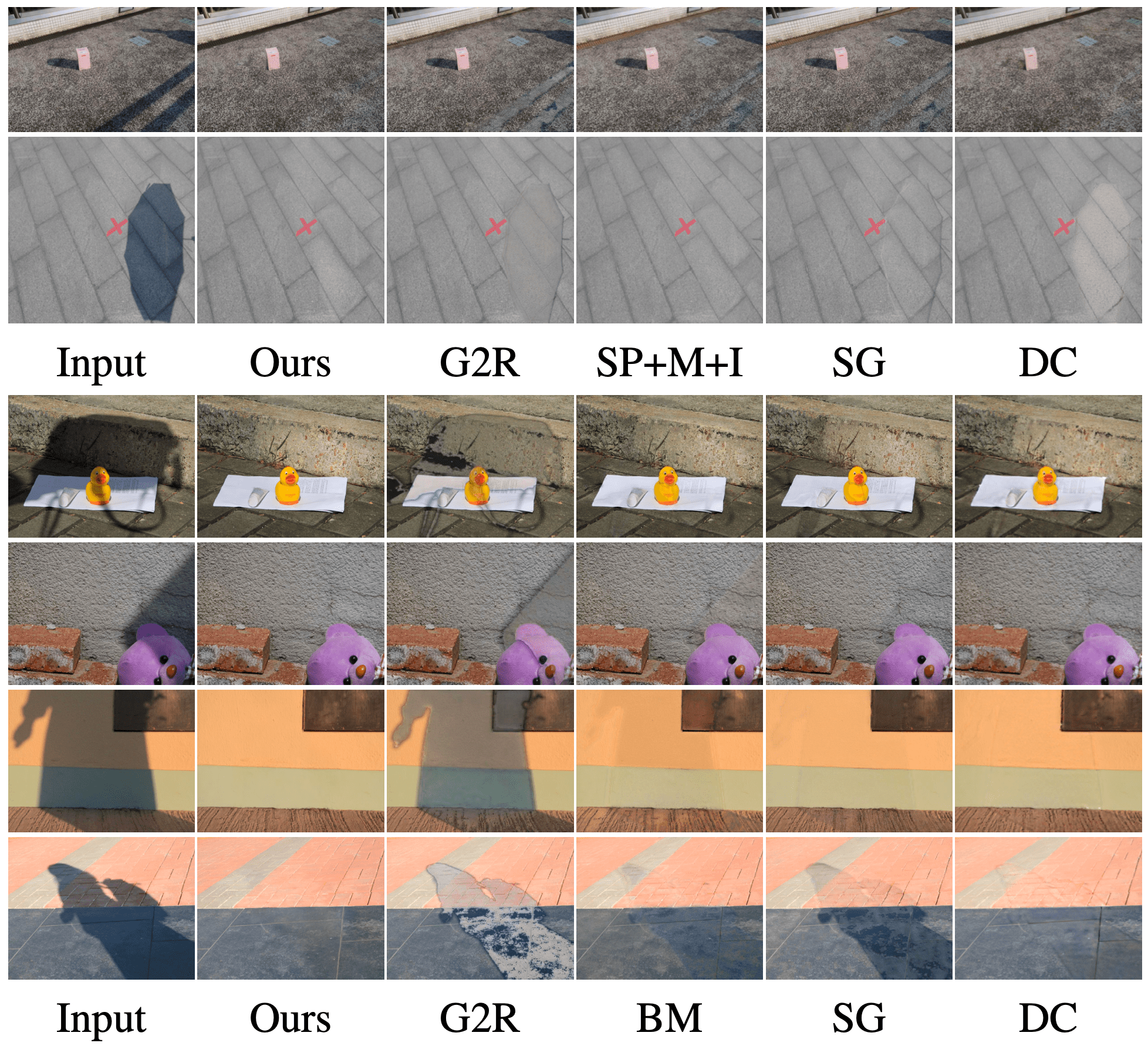}}
	\caption{Hard shadow removal results on the USR (top), AISTD (second) and SRD (bottom four rows) datasets.}
	\label{fig:hard}
\end{figure}

\begin{figure}[t]
	\centering
	\captionsetup[subfigure]{font=small, labelformat=empty}
	{\includegraphics[width=8.2cm]{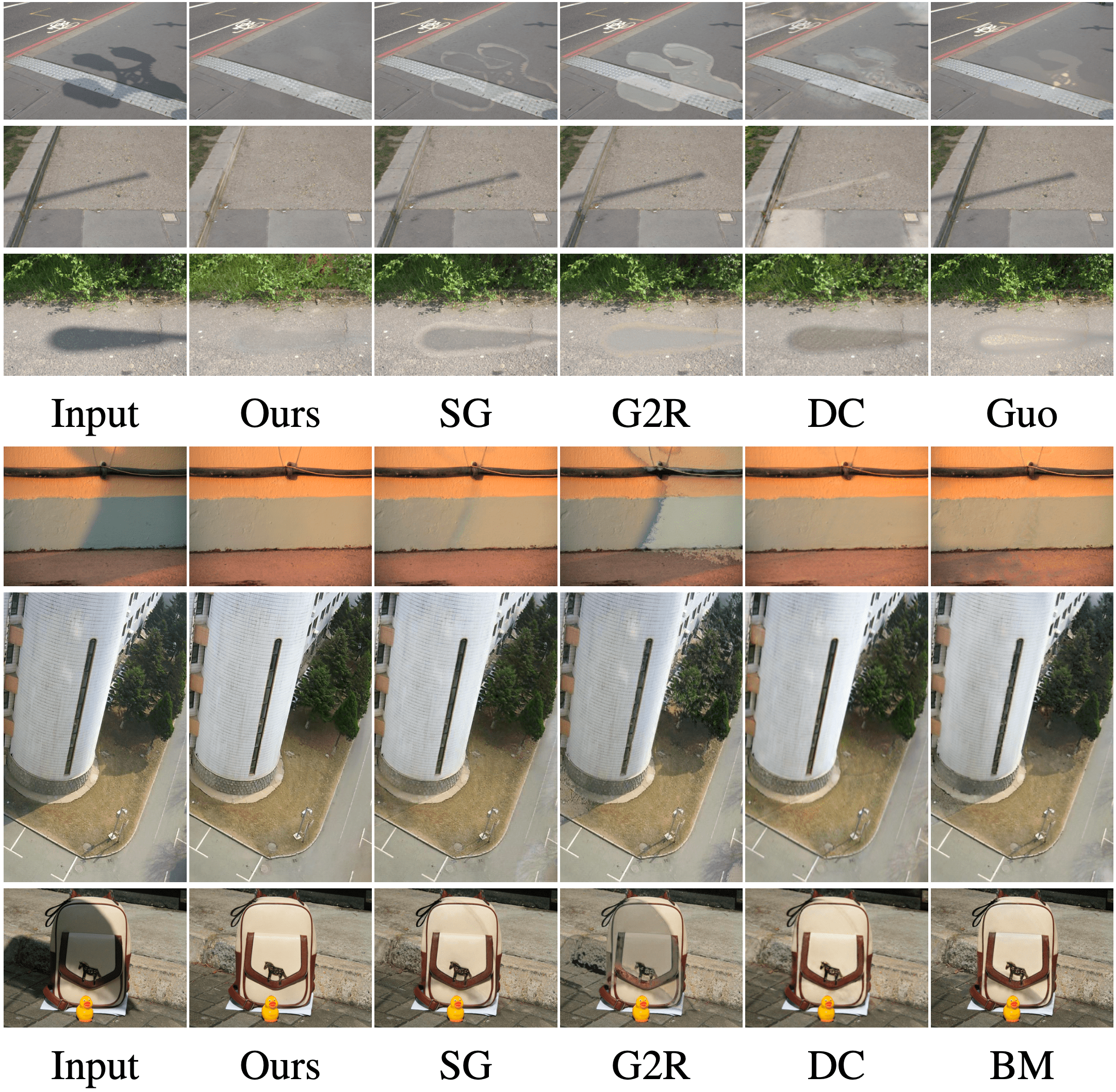}}
	\caption{Shadow removal results on the soft shadow dataset LRSS (top three) and SRD (bottom three rows) datasets. Note that our method does not need additional masks in training and testing.}
	\label{fig:soft}
\end{figure}

\paragraph{Implementation}
To ensure fair comparisons, all the baselines, including ours are trained and tested on the same datasets. 
We trained our DeS3 on each dataset and tested on the corresponding dataset, 
e.g., for SRD~\cite{qu2017deshadownet}, we used 2680 SRD and 408 SRD images for training and testing, respectively. 
We use 1000 steps for training and 25 steps for inference, noise schedule $\beta_t$ linear from 0.0001 to 0.02. 
$\alpha$ and $\beta$ are empirically set to 0.5 in training.
$\mathcal{L}_{\text t} (\xw, \x) = \lambda_{\rm sim} \mathcal{L}_{\rm sim}$, which is the total reverse sampling loss, where $\lambda_\text{sim}=2$, we balance the self-tuned ViT loss to select the good sampling output.

\paragraph{Shadow Removal on Hard Shadows} 
The SRD dataset consists of 2680 training and 408 testing pairs of shadow and shadow-free images without shadow masks. It contains hard shadows and soft shadows. 
For the baseline methods that require shadow masks, we additionally use the processing results of DHAN~\cite{cun2020towards} for SRD shadow masks.

Fig.~\ref{fig:hard} (bottom four rows) and Table~\ref{tb:srd} show the hard shadow removal results on the SRD dataset.
Fig.~\ref{fig:soft} (bottom three rows) show the soft shadow removal results on the SRD dataset, which show our DeS3 outperforms the SOTA methods.
The best values for each metric are highlighted in \textbf{bold}.
Moreover, DeS3 achieves the highest PSNR and SSIM values in shadow, non-shadow and overall areas. PSNR and SSIM are widely used in image generation tasks~\shortcite{wang2023seeing,wang2023promptrestorer,wang2023selfpromer}.
Fig.~\ref{fig:hard} (second row) show results on the AISTD dataset, which demonstrate the superiority of our DeS3 compared to baseline results.
The USR dataset~\cite{hu2019mask} is an unpaired dataset, including hard, soft, and self shadows. 
We evaluate on USR test shadow images and show results in Fig.~\ref{fig:hard} (first row).

\begin{table*}[t]
	\centering
	\renewcommand{\arraystretch}{1.2}
	\resizebox{0.71\textwidth}{!}{
		\begin{tabular}{c|c|cccc|cc|cc}
			\toprule
			Method   &Guo   &Gryka &DHAN &SP+M &MS.GAN &DC &Guo (at) & S.Diff. &Ours\\\hline
			RMSE$\downarrow$   &6.02     &4.38       &7.92 &7.48   &7.13     &\underline{3.48}  &5.87 &3.49 &\bf{3.01}\\
			PSNR$\uparrow$        &27.88    &29.25      &25.57&23.93  &25.12    &\underline{31.01} &28.02 &28.86 &\bf{33.95} \\\hline
			Train     &P+M           &P+M+S      &P+M+S      &P+M  &UP &UP &P  &P+M &P\\
			\bottomrule
	\end{tabular}}
	\caption{Results on the LRSS (soft shadow) dataset. M and S represent ground truth shadow masks and synthetic data are used in training. ``P" and ``UP" stand for ``Paired" and ``Unpaired". Our DeS3 does not need shadow masks.}
	\label{tb:lrss}
\end{table*}

\begin{figure}[t!]
	\centering
	\setcounter{subfigure}{0}
	{\includegraphics[width=8.2cm]{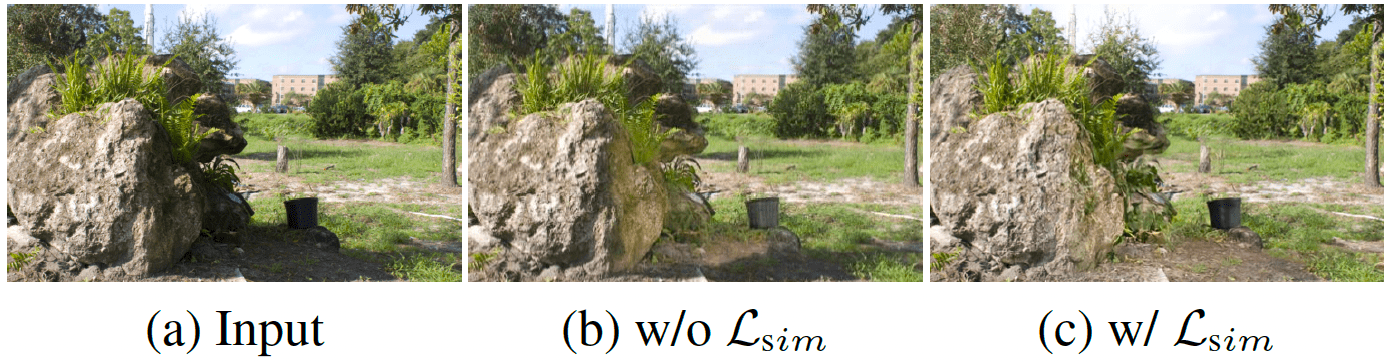}}
	\caption{Ablation studies on the ViT similarity loss.}
	\label{fig:ab}
\end{figure}

\paragraph{Shadow Removal on Soft and Self Shadows} 
The LRSS~\cite{gryka2015learning} is a soft shadow dataset with 134 shadow images\footnote{LRSS Dataset are obtained from their project website: http://visual.cs.ucl.ac.uk/pubs/softshadows/}; we followed~\cite{jin2021dc,gryka2015learning}, using the same 34 LRSS images with their corresponding shadow-free images for evaluation shown in Table~\ref{tb:lrss}. Our DeS3 achieves the highest PSNR and lowest RMSE.

Fig.~\ref{fig:soft} (top three rows) show results on the LRSS dataset. 
Besides the SOTA baselines, we compared our method with the traditional hard and soft shadow removal method~\cite{guo2012paired}.
Since the majority of the SOTA baselines need the shadow mask for evaluation, we use the shadow mask provided by LRSS dataset~\cite{gryka2015learning}.
However, the provided binary shadow mask does not consider the ambiguous boundaries of soft and self shadows, thus leading to remaining shadows in the ambiguous boundaries.

UCF~\cite{zhu2010learning} and UIUC dataset~\cite{guo2011single}, contain 245 and 108 images. 
UIUC provides 30 images where shadows are caused by objects in the scene, which are used for self shadow evaluation,
the results are shown in Fig.~\ref{fig:self}.
For the baseline methods that require shadow masks, we additionally use the detection results of BDRAR~\cite{zhu2018bidirectional}.

\begin{table}[t!]
	\centering
	\renewcommand{\arraystretch}{1.2}
	\resizebox{0.45\textwidth}{!}{	
		\begin{tabularx}{\columnwidth}{ c|Y|Y|Y }
			\toprule
			Method					   &S $\downarrow$ &NS $\downarrow$ &Overall$\downarrow$ \\\hline
			w/o  Adaptive Classifier Atten &6.35 &3.02 &4.46\\\hline
			w/o $\mathcal{L}_{\text sim}$&6.21 &2.99 &3.94\\\hline
			Ours (complete model)      &\bf 5.88 &\bf 2.83  &\bf 3.72\\
			\bottomrule	
	\end{tabularx}}
	\caption{Ablation experiments of our method using the SRD dataset. The numbers represent RMSE, lower is better.}
	\label{tb:ablation}
\end{table}

\paragraph{Ablation Studies}
Fig.~\ref{fig:ab} and Table.~\ref{tb:ablation} show the effectiveness of the ViT similarity loss used in DeS3.
From Fig.~\ref{fig:ab} (second row), the ViT similarity loss help in maintaining the structures of the mountain, owing to the features (the self-similarity keys) of the pre-trained DINO-ViT.
We compare our method with two diffusion baselines: WeatherDiffusion~\cite{ozdenizci2023restoring} and SR3~\cite{saharia2022image} shown in Tab.\ref{tb:srd}.
More results are in the supplementary material.
\section{Conclusion}
\label{sec:conclusion}
We have proposed DeS3, the first shadow removal network that performs shadow removal robustly on hard, soft and self shadows.
Unlike existing methods, our DeS3 does not rely on masks during training and testing.
To guide our reverse sampling process to preserve object/scene structure information, we propose the self-tuned ViT similarity loss.
Unlike existing methods, our DeS3 performs shadow removal robustly on outdoor self shadows, which is intractable to have ground truths.

\section*{Acknowledgments}
This research is supported by the National Research Foundation, Singapore under its AI Singapore Programme (AISG Award No: AISG2-PhD/2022-01-037[T]). 
Wenhan Yang's research is supported in part by the Basic and Frontier Research Project of PCL, and the Major Key Project of PCL.

\clearpage
\bibliography{aaai24}

\end{document}